\documentclass[10pt,twocolumn,letterpaper]{article}

\usepackage{cvpr}

\usepackage{graphicx}
\usepackage{amsmath}
\usepackage{amssymb}
\usepackage{booktabs}
\usepackage{multirow}
\usepackage{bbding} 
\usepackage{enumerate}
\usepackage{float}
\usepackage{bm}
\usepackage{balance}

\usepackage[pagebackref,breaklinks,colorlinks]{hyperref}

\usepackage[capitalize]{cleveref}
\crefname{section}{Sec.}{Secs.}
\Crefname{section}{Section}{Sections}
\Crefname{table}{Table}{Tables}
\crefname{table}{Tab.}{Tabs.}

\def\cvprPaperID{5992}
\def\confName{CVPR}
\def\confYear{2022}

\begin{document}

\title{Adaptive Trajectory Prediction via Transferable GNN}

\author{Yi Xu\textsuperscript{1}, Lichen Wang\textsuperscript{1}, Yizhou Wang\textsuperscript{1}, Yun Fu\textsuperscript{1,2} \\
\textsuperscript{1}Department of Electrical and Computer Engineering, Northeastern University, USA\\
\textsuperscript{2}Khoury College of Computer Science, Northeastern University, USA\\
{\tt\small xu.yi@northeastern.edu, \{wanglichenxj,wyzjack990122\}@gmail.com, yunfu@ece.neu.edu}}
\maketitle

%%%%%%%%% ABSTRACT
\begin{abstract}
Pedestrian trajectory prediction is an essential component in a wide range of AI applications such as autonomous driving and robotics. Existing methods usually assume the training and testing motions follow the same pattern while ignoring the potential distribution differences (e.g., shopping mall and street). This issue results in inevitable performance decrease. To address this issue, we propose a novel Transferable Graph Neural Network (T-GNN) framework, which jointly conducts trajectory prediction as well as domain alignment in a unified framework. Specifically, a domain-invariant GNN is proposed to explore the structural motion knowledge where the domain-specific knowledge is reduced. Moreover, an attention-based adaptive knowledge learning module is further proposed to explore fine-grained individual-level feature representations for knowledge transfer. By this way, disparities across different trajectory domains will be better alleviated. More challenging while practical trajectory prediction experiments are designed, and the experimental results verify the superior performance of our proposed model. To the best of our knowledge, our work is the pioneer which fills the gap in benchmarks and techniques for practical pedestrian trajectory prediction across different domains.
\end{abstract}

%%%%%%%%% BODY TEXT
\section{Introduction}\label{sec:intro}
Trajectory prediction aims to predict the future trajectory seconds to even a minute prior from a given trajectory history. It plays an indispensable role in a large number of real world applications such as autonomous driving, robotics, navigation, video surveillance, and so on. In self-driving scenario, accurate pedestrian trajectory prediction is essential for planning~\cite{Bai2015Intention, ma2020optimal}, decision making~\cite{Yuanfu2018PORCA}, environmental perception~\cite{Talbot2020Robot,Obo2020Intelligent}, person identification~\cite{Luber2010People}, and anomaly detection~\cite{Musleh2010Identifying,2004Pedestrian}. Trajectory prediction is a challenging task. For instance, strangers tend to walk alone trying to avoid collisions but friends tend to walk as a group~\cite{Moussaid2010TheWalking}. In addition, pedestrians can interact with surrounding objects or other pedestrians, while such interaction is too complex and subtle to quantify. To consider such interactions, a pooling layer is designed in work Social-LSTM~\cite{alahi2016social} to pass the interaction information among pedestrians, and then a long short-term memory (LSTM) network is applied to predict future trajectories. Following this pattern, many methods~\cite{liang2019peeking, zhang2019sr, hu2020collaborative, xu2020cf, zhu2021simultaneous} have been proposed for sharing information via different mechanisms, i.e., attention mechanism or similarity measure. Instead of predicting one determined future trajectory, some generative adversarial network-based (GAN)~\cite{fernando2018gd, gupta2018social, li2019conditional, sadeghian2019sophie, dendorfer2021mg} and encoder-decoder-based methods~\cite{Mangalam2020It, cheng2020exploring, salzmann2020trajectronplusplus, xu2021tra2tra, chen2021personalized, chen2021human, shafiee2021introvert} have been proposed to generate multiple feasible trajectories.

\begin{figure}[t]
	\begin{center}
		\includegraphics[width=0.45\textwidth]{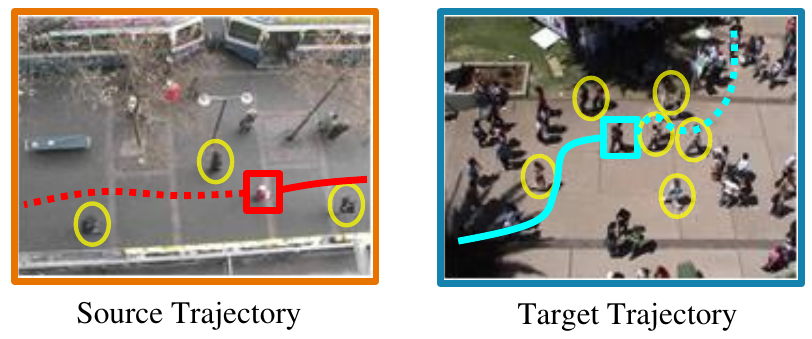} 
	\end{center}
	\vspace{-5mm}
	\caption{An example that reveals the limitation of original learning strategy. These two frames are extracted from two different scenes and there is a huge difference between these trajectories.}\label{intro}
	\vspace{-1mm}
\end{figure}

\begin{table}[t]
	%\vspace{1mm}
	\centering
	\scalebox{0.66}{
	\begin{tabular}{c|ccccc|cc}
		\toprule
		\multirow{2}{*}{Metric} &\multicolumn{5}{c|}{Trajectory Domains} & \multirow{2}{*}{\textbf{\textit{E-D}}} & \multirow{2}{*}{\textbf{\textit{S-D}}} \\
		\cmidrule{2-6}
		& ETH & HOTEL & UNIV & ZARA1 & ZARA2 &  &  \\
		\midrule
		NoS & \textbf{\color{blue}70} & 301 & \textbf{\color{red}947} & 602 & 921 & 877 & 383.63\\
        NoP & \textbf{\color{blue}181} & 1053 & \textbf{\color{red}24334} & 2253 & 5833 & 24153 & 10073.07\\
        AN & \textbf{\color{blue}2.586} & 3.498 & \textbf{\color{red}25.696} & 3.743 & 6.333 & 23.11 & 9.78\\
        AV ($m/s$)& \textbf{\color{red}0.437} & \textbf{\color{blue}0.178} & 0.205 & 0.369 & 0.206 & 0.259 & 0.11 \\
        AA ($m/s^2$)& \textbf{\color{red}0.131} & 0.06 & 0.035 & 0.039 & \textbf{\color{blue}0.026} & 0.105 & 0.04\\
		\bottomrule
	\end{tabular}
	}
	\vspace{-1mm}
	\caption{Statistics of five different scenes, ETH, HOTEL, UNIV, ZARA1, and ZARA2. NoS denotes the number of sequences to be predicted, NoP denotes the number of pedestrians, AN denotes the average number of pedestrians in each sequence, AV denotes the average velocity of pedestrians in each sequence, and AA denotes the average acceleration of pedestrians in each sequence. \textbf{\textit{E-D}} represents Extreme Deviation and \textbf{\textit{S-D}} represents Standard Deviation.}
	\label{intro:sta}
	\vspace{-1mm}
\end{table}

However, these existing methods usually focus on learning a generic motion pattern while ignoring the potential distribution differences between the training and testing samples. We argue that this learning strategy has some limitations. \cref{intro} illustrates one basic concept. It is obvious that the trajectories of walking pedestrians in different trajectory domains are different, the trajectory in the left figure is stable but the trajectory in the right figure is much more tortuous. The original strategy is to learn these two samples together without considering distribution differences, which introduces domain-bias and disparities into the model.

In order to quantitatively and objectively evaluate the potential domain gaps, \cref{intro:sta} gives five numerical statistics of five commonly used trajectory domains. We can observe that the number of pedestrians in UNIV is much larger than that in ETH, and the differences among five trajectory domains are significant. As for pedestrian moving pattern, pedestrians in ETH have the largest average moving velocity, which is nearly three times larger than that in HOTEL. In addition, pedestrians in ETH also have the largest average moving acceleration, which is nearly five times larger than that in ZARA2. The E-D value and S-D value also reveal the huge differences among five different trajectory domains. This situation is general and always exists in practical applications. For example, in vision applications, cameras located in different cities/corners could lead to significant distribution gap. Similar situations are also common in robot navigation or autonomous driving-related applications since the environments are constantly changing.

\begin{figure}[t]
	\begin{center}
		\includegraphics[width=0.45\textwidth]{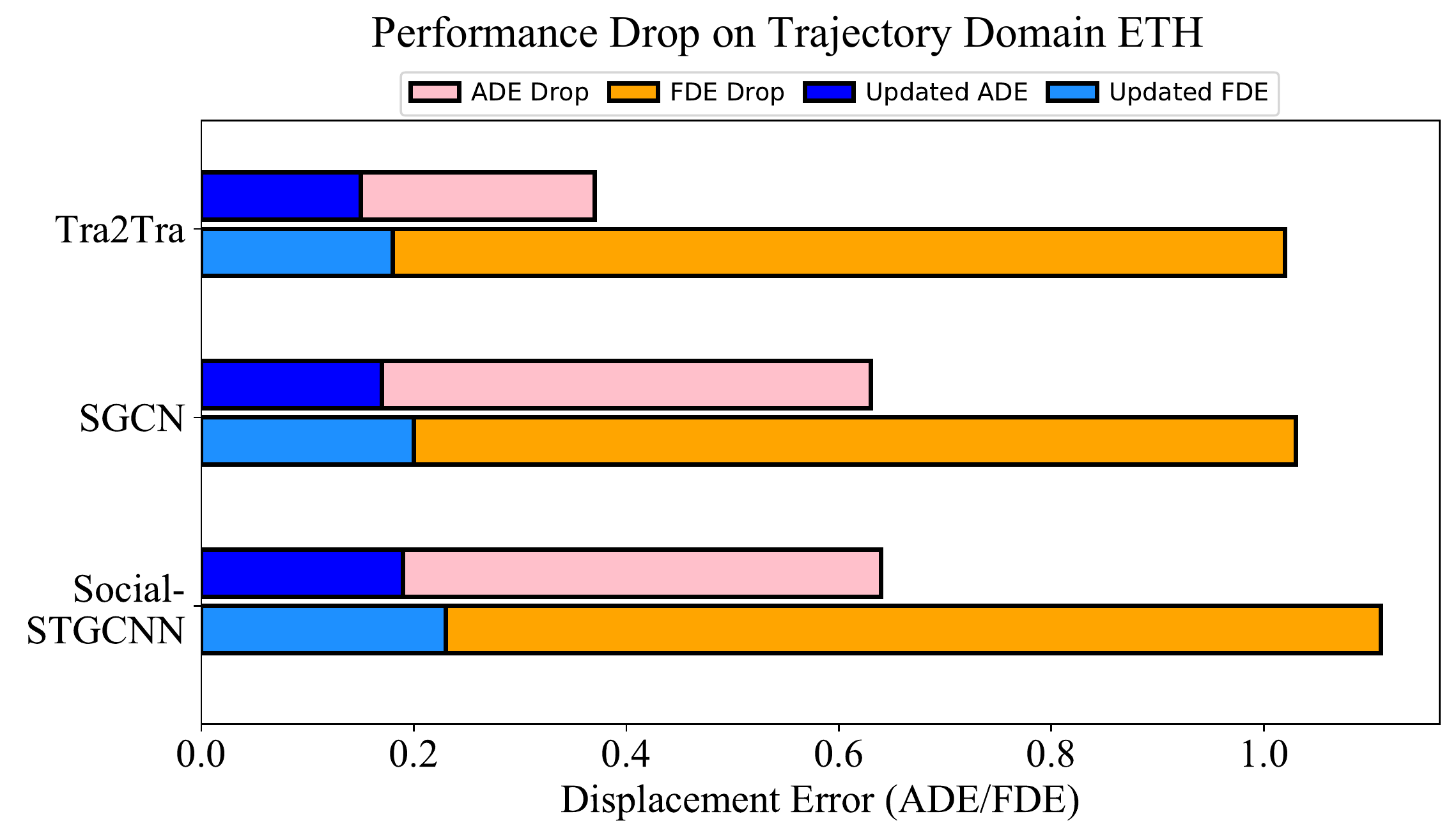}
	\end{center}
	\vspace{-5mm}
	\caption{Performance comparison of three state-of-the-art methods under the original leave-one-out setting and the consistent setting. The performance drops of all three models are significant.}\label{drop}
	\vspace{-2mm}
\end{figure}

To further demonstrate this challenge, we apply three state-of-the-art methods, Social-STGCNN~\cite{mohamed2020social}, SGCN~\cite{shi2021sgcn}, Tra2Tra~\cite{xu2021tra2tra} to demonstrate the performance drop when it comes to different trajectory domains. We take ETH as the example, these models are trained on the validation set of ETH and evaluated on the standard testing set of ETH. Note that there is no overlap trajectory sample between the training and testing set, but the distributions of them can be regarded as consistent. We refer to this evaluation setting as ``consistent setting'' and the performance under this new protocol as ``updated ADE'' and ``updated FDE''. \cref{drop} shows the updated ADE/FDE as well as the original ADE/FDE reported in their papers. The performance drops are significant which further reveal the domain-bias problem in the original leave-one-out setting.

Domain adaptation (DA) is a subcategory of transfer learning which aims to address the domain shift issue. The basic idea is to minimize the distance of distributions of source and target domains via some distance measures, such as maximum mean discrepancy (MMD)~\cite{Ni2013Subspace,Long2015Learning}, correlation alignment distance (CORAL)~\cite{Sun2016Deep,Zhuo2017Deep}, and adversarial loss~\cite{Ganin2014Unsupervised,2020Unsupervised}. Among these methods, the feature dimension of one sample is fixed in both source and target domain. On the contrary, a ``sample'' in our task is a combination of multiple trajectories with different pedestrians, which has not only global domain shift but also internal correlations. Therefore, directly utilizing the general feature representation of one ``sample'' results in the lack of crucial individual-level fine-grained features. Consequently, the most popular domain adaptation approaches are not applicable here.

In this work, we delve into the trajectory domain shift problem and propose a transferable graph neural network via adaptive knowledge learning. Specifically, we propose a novel attention-based adaptive knowledge learning module for trajectory-to-trajectory domain adaptation. In addition, a novel trajectory graph neural network is presented. It is able to extract comprehensive spatial-temporal features of pedestrians that enhance the domain-invariant knowledge learning. The contributions of our work are summarized as,

\begin{itemize}
\vspace{-1mm}
\item We delve into the domain shift problem across different trajectory domains and propose a unified T-GNN method for jointly predicting future trajectories and adaptively learning domain-invariant knowledge. 
\vspace{-1mm}
\item We propose a specifically designed graph neural network for extracting comprehensive spatial-temporal feature representations. We also develop an effective attention-based adaptive knowledge learning module to explore fine-grained individual-level transferable feature representations for domain adaptation.
\vspace{-1mm}
\item We introduce a brand new setting for pedestrian trajectory prediction problem, which is meaningful in real practice. We set up strong baselines for pedestrian trajectory prediction under this domain-shift setting.
\vspace{-1mm}
\item Experiments on five trajectory domains verify the consistent and superior performance of our method.
\vspace{-1mm}
\end{itemize}

As it is natural to use a graph-based model to represent the topology of social networks, recent methods~\cite{mohamed2020social, ivanovic2019trajectron, sun2020recursive, li2021spatial, shi2021sgcn, wang2021graphtcn} employ graph neural networks as their backbones. Different from these methods, the graph neural network we employed is simple yet specifically designed not only to extract effective spatial-temporal features but also to be suitable for domain-invariant knowledge learning. 

% ===================
% Related works
% ===================
\section{Related Works}
\begin{figure*}[ht]
	\centering
	\includegraphics[width=1.0\textwidth]{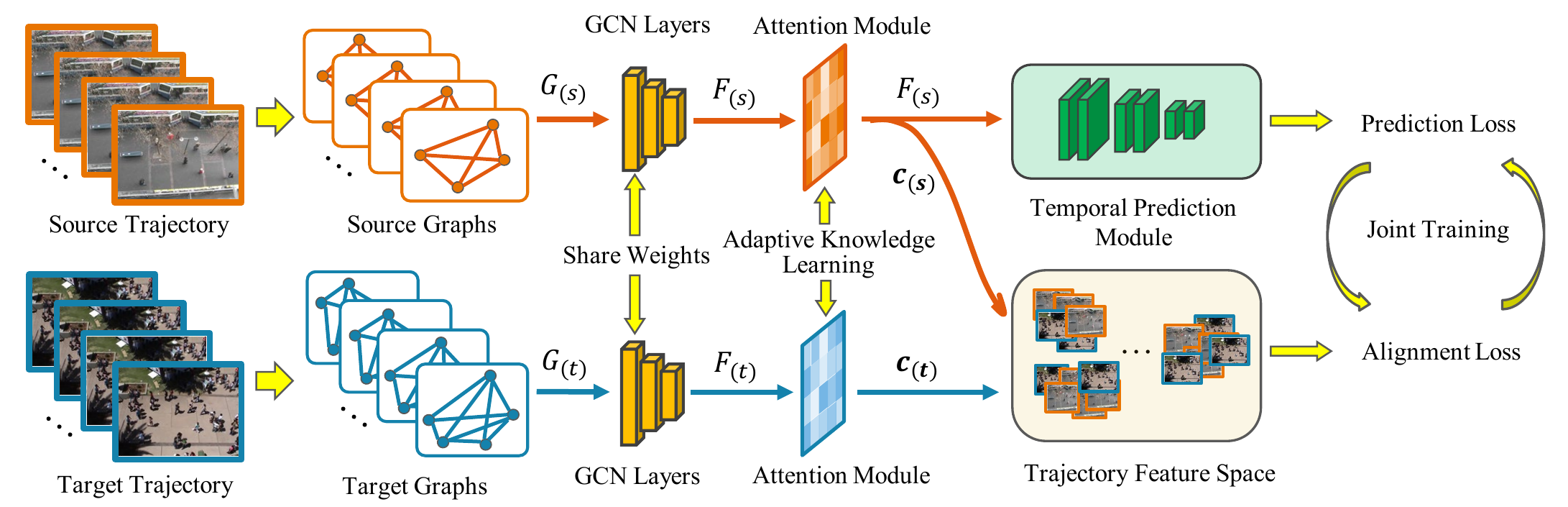} 
	\caption{Flowchart of our T-GNN model. Given the source and target trajectories, we first construct corresponding successive graphs $G_{(s)}$ and $G_{(t)}$, and then GCN layers are applied to extract feature representations $F_{(s)}$ and $F_{(t)}$ from these graphs. Following this, $F_{(s)}$ and $F_{(t)}$ are forwarded through the Attention-Based Adaptive Knowledge Module to learn transferable features $\textbf{c}_{(s)}$ and $\textbf{c}_{(t)}$ for aligning the source and target trajectory domain. Afterwards, only $F_{(s)}$ from source trajectory domain is utilized for future trajectory prediction via Temporal Prediction Module. Finally, our T-GNN model jointly minimizes the prediction loss and alignment loss.}
	\label{framework}
\end{figure*}

% Forecasting pedestrian trajectory
\subsection{Forecasting Pedestrian Trajectory}
Forecasting pedestrian trajectory aims to predict future locations of the target person based on his/her past locations and surroundings. Early researches attempt to use mathematical models~\cite{crowsourcing_1} to make predictions such as Gaussian Process~\cite{Keat2007Modelling, Ellis2009Modelling}, and Markov Decision Process~\cite{Makris2002Spatial, Kitani2012Activity}. 
Recently, a large number of deep learning methods have been proposed to solve this prediction problem. In the work  Social-LSTM~\cite{alahi2016social}, pedestrians are modeled with Recurrent Neural Networks (RNNs), and the hidden states of pedestrians are integrated via a designed pooling layer, where human-human interaction features are shared. 
To improve the quality of extracted interaction features, many recent works~\cite{vemula2018social, zhang2019sr, liang2019peeking, bisagno2018group, hu2020collaborative, zheng2021unlimited} follow this idea to pass information among pedestrians, and different effective message passing approaches are proposed. Taking into account the uncertainty of pedestrians walking, some studies~\cite{sadeghian2019sophie, li2019conditional, vaswani2017attention,fernando2018gd, amirian2019social, kosaraju2019social, dendorfer2021mg} utilize Generative Adversarial Networks (GAN) to make multiple plausible predictions of each person. In addition, different Encoder-Decoder structures~\cite{Mangalam2020It,cheng2020exploring, sun2021three} are also applied in this task, which are more flexible to encode different useful context features. 

Transformer structure~\cite{vaswani2017attention} has achieved remarkable performance in Natural Language Processing field~\cite{Devlin2018BERT}. Motivated by this design, some studies~\cite{giuliari2020transformer, yu2020spatio, yuan2021agent} adopt it to the trajectory prediction task and improve the overall prediction precision. For the past two years, some works~\cite{tran2021goal, zhao2021you, mangalam2021goals} have been proposed to explore the goal-driven trajectory prediction. The main idea is to estimate the end points of trajectories for prediction guidance. In addition, some interesting perspectives have been introduced into this task, i.e., long-tail situation~\cite{makansi2021exposing}, energy-based model~\cite{pang2021trajectory}, interpretable forecasting model~\cite{kothari2021interpretable}, active-learning~\cite{xu2021robust}, and counterfactual analysis~\cite{chen2021human}. Different from recent work~\cite{liang2020simaug} that studies the problem of predicting future trajectories in unseen cameras with only 3D simulation data, our work is carried out under a more general and practical trajectory prediction setting, which has more profound influences.

% GNN-based trajectory prediction models
\subsection{Graph-Involved Forecasting Models}
Thanks to the powerful representation ability in non-Euclidean space, Graph Neural Networks (GNNs) are widely applied in the trajectory prediction task~\cite{velivckovic2017graph, yan2018spatial, wu2020comprehensive, jain2016structural, wang2019inductive} recently. The basic idea is to treat the pedestrians as the nodes in a graph while measuring their interactions via graph edges. Recent works have utilized different variants of graph neural networks, e.g., edge-feature aggregation~\cite{Rosmann2017Online, sun2020recursive}, spatial-temporal feature extraction~\cite{mohamed2020social, ivanovic2019trajectron}, adapted graph structure~\cite{VectorNet2020, zhu2019starnet, mohamed2020social, shi2021sgcn}, and graph attention method~\cite{kosaraju2019social}. Our work also applies the graph model for feature representations extraction. Different from the above methods, our model is specifically designed for effective spatial-temporal feature representation learning as well as trajectory domain-invariant knowledge learning.

% Domain adaptation
\subsection{Domain Adaptation}
Recently, domain adaptation (DA) problem has attracted considerable attention, motivating a large number of approaches~\cite{Yang2018Learning,Ding2018Graph} to resolve the domain shift problem. Generally speaking, it can be divided into two main categories, one is semi-supervised DA problem, and the other is unsupervised DA problem. The difference between these two categories lies in the accessibility of target labels in the training phase. In semi-supervised DA~\cite{Hal2010Co,Saenko2010Adapting,He2020Classification}, only a small number labeled target samples is accessible. 

In unsupervised DA~\cite{2015Geodesic,Luo2018Deep,Jim2020A,cai2021graph}, the target domain is totally unlabeled, which is much more challenging. In our work, we are dealing with the unsupervised DA problem. The majority of existing unsupervised DA methods usually project the source and target samples into a shared feature space, and then align their feature distributions via minimizing some distance measures, such as MMD~\cite{Ni2013Subspace,Long2015Learning}, CORAL~\cite{Sun2016Deep,Zhuo2017Deep}, or Adversarial Loss~\cite{Ganin2014Unsupervised,2020Unsupervised} to force their distributions indistinguishable. As discussed above, these methods cannot be directly applied in our work. We address this problem by introducing an attention-based adaptive knowledge learning module for knowledge transfer.

% ===================
% Our method
% ===================
\section{Our Method}
The overall framework of T-GNN model is illustrated in~\cref{framework}. It consists of three main components: 1) a graph neural network to extract effective spatial-temporal features of pedestrians from both source and target trajectory domains, 2) an attention-based adaptive knowledge learning module to explore domain-invariant individual-level representations for transfer learning, 3) a temporal prediction module for future pedestrian trajectory predictions.

\subsection{Problem Definition}
Given one pedestrian $i$ observed trajectory $\Gamma^i=\{o^{i}_{1},...,o^{i}_{obs}\}$ from time step $T_1$ to $T_{obs}$, aim to predict the future trajectory $\overline{\Gamma^i}=\{o^{i}_{obs+1},...,o^{i}_{pred}\}$ from time step $T_{obs+1}$ to $T_{pred}$, where $o^{i}_{t}=(x^{i}_{t},y^{i}_{t})\in \mathbb{R}^2$ denote the coordinates. Considering all the pedestrians in the scene, the goal is to predict trajectories of all the pedestrians simultaneously by a model $f(\cdot)$ with parameter $W^{*}$. Formally,
\begin{equation}
\overline{\Gamma} = f(\Gamma^1,\Gamma^2,...,\Gamma^N;W^{*}),
\label{}
\end{equation}
where $\overline{\Gamma}$ is the set of future trajectories of all the pedestrians, $N$ denotes the number of pedestrians, and $W^{*}$ represents the collection of learnable parameters in the model.

\subsection{Spatial-Temporal Feature Representations}
Different from traditional time series forecasting, it is more challenging to predict pedestrian future trajectories because of the implicit human-human interactions and their strong temporal correlations. Therefore, extracting comprehensive spatial-temporal feature representations of observed pedestrian trajectories becomes a key point to accurately predict trajectories. In our work, considering the data structure of trajectories, a graph neural network is first employed to extract spatial-temporal feature representations.

Before constructing the graph, coordinates of all pedestrians are firstly passed through one layer as,
\begin{equation}
    {o'}^{i}_{t}=o^{i}_{t}-\frac{1}{N}\sum_{i=1}^{N}o^{i}_{obs},
    \label{dc}
\end{equation}
where $N$ is the number of pedestrians in the scene, $o^{i}_{obs}$ represents the coordinates of pedestrian $i$ at the last observed frame $T_{obs}$. This decentralization operation is able to eliminate the effects of scene size differences and is also applied in recent works~\cite{zhu2019starnet, xu2021tra2tra}. We refer to ${o'}^{i}_{t}=({x'}^{i}_{t},{y'}^{i}_{t})$ as the ``relative coordinates'' for the following graph construction.

We define the graph $G_{t}=(V_{t},E_{t},F_{t})$, where $V_{t}=\{v_{t;i} |i=1,...,N \}$ is the vertex set of pedestrians in the graph, $E_{t}=\{e_{t;i,j}|i,j=1,...,N\}$ is the edge set that indicates the relationship between two pedestrians, and $F_{t}=\{f_{t;i} |i=1,...,N \}\in \mathbb{R}^{N \times D_{f}}$ is the feature matrix associated with each pedestrian $v_{t;i}$ ($D_{f}$ is the feature dimension). The topological structure of graph $G_{t}$ is represented by the adjacency matrix $A_{t}=\{a_{t;i,j}|i,j=1,...,N\}\in \mathbb{R}^{N \times N}$. In our case, the value of $a_{t;i,j}$ in adjacency matrix $A_t$ is initialized as the distance between pedestrian $i$ and $j$ as,
\begin{equation}
    a_{t;i,j}= \|{o'}^{i}_{t} - {o'}^{j}_{t} \|_{2},
    \label{eq:a_value}
\end{equation}
where $\|*\|_2$ is the $L_2$ distance, and ${o'}^{i}_{t}$ denotes the ``relative coordinates'' ${o'}^{i}_{t}=({x'}^{i}_{t},{y'}^{i}_{t})$ of pedestrian $i$ at time step $t$. As it should be other possible definitions of $a_{t;i,j}$, we additionally investigate and analysis other three different definitions of $a_{t;i,j}$, and results indicate that using $L_2$ distance is more appropriate in this situation.

The value of $f_{t;i}$ in feature matrix $F_t$ is defined as,
\begin{equation}
    f_{t;i}=\sigma(({x'}^{i}_{t},{y'}^{i}_{t});\textbf{W}_{o}),
    \label{dcl}
\end{equation}
where $\textbf{W}_{o}\in \mathbb{R}^{2 \times D_{f}}$ are projection learnable parameters, $\sigma(\cdot)$ is $\operatorname{ReLU}$ non-linearity activation function.

To measure the relative importance of dynamic spatial relations between pedestrians, the graph attention layer from~\cite{velivckovic2017graph} is adopted here to update the adjacency matrix $A_{t}$. The graph attention coefficients are calculated as,
\begin{equation}
\alpha_{t;i,j}=\frac{\exp\left(\phi\left(\textbf{W}_{l}\left[\textbf{a}_{t;i}\oplus \textbf{a}_{t;j}\right]\right)\right)}{\sum_{j=1}^{N}\exp\left(\phi\left(\textbf{W}_{l}\left[\textbf{a}_{t;i}\oplus \textbf{a}_{t;j}\right]\right)\right)},
\label{eq1}
\end{equation}
where $\textbf{a}_{t;i} \in \mathbb{R}^{N\times 1} $ is $i^{th}$ column vector in $A_{t}$, $\textbf{W}_{l} \in \mathbb{R}^{1 \times 2N}$ are learnable parameters, $\oplus$ represents the concatenation that operates in the dimension of row, $\phi$ is $\operatorname{LeakyReLU}$ non-linearity activation function with $\theta=0.2$. The same parameters are used here, see~\cite{velivckovic2017graph} for details.

The linear combination $\textbf{p}_{t;i}$ is thus computed according to the obtained attention coefficients. Formally, we have,
\begin{equation}
\textbf{p}_{t;i}=\sigma\left(\sum_{j=1}^{N}\alpha_{t;i,j}\textbf{a}_{t;j}\right).
\label{eq2}
\end{equation}

With each column vector $\textbf{p}_{t;i}$ concatenated together, we obtain the new updated adjacency matrix $A_{t}' \in \mathbb{R}^{N\times N}$, which contains the information of global spatial features of pedestrians at time step $t$. Then, the GCN layers~\cite{kipf2016semi} are applied here to further extract spatial-temporal features. Similar with~\cite{mohamed2020social}, we first add identity matrix to $\hat{A}_{t}$ as,
\begin{equation}
\hat{A}_{t}=A_{t}'+I.
\label{eq3}
\end{equation}

Then, we stack $\hat{A}_{t}$ from time step $T_1$ to $T_{obs}$ as $\hat{A}=\{\hat{A}_{1},\hat{A}_{2},...,\hat{A}_{obs}\} \in \mathbb{R}^{N\times N\times L_{obs}}$ and also stack vertex feature matrices of the $l^{th}$ layer from time step $T_1$ to $T_{obs}$ as $F_{t}^{(l)}=\{F_{1}^{(l)},F_{2}^{(l)},...,F_{obs}^{(l)}\}\in \mathbb{R}^{N\times D_{f}\times L_{obs}}$, where $L_{obs}$ represents the observation length. In addition, the stack of node degree matrices $D=\{D_{1},D_{2},...,D_{obs}\}$ are correspondingly calculated from $\{\hat{A}_{1},\hat{A}_{2},...,\hat{A}_{obs}\}$.

Finally, the output $F^{(l+1)}\in \mathbb{R}^{N\times D_{f}\times L_{obs}}$ of the $(l+1)^{th}$ layer is calculated as,
\begin{equation}
F^{(l+1)}=\sigma\left(D^{-\frac{1}{2}}\hat{A}D^{\frac{1}{2}}F^{(l)}\textbf{W}^{(l)}\right),
\label{eq4}
\end{equation}
where $\textbf{W}^{(l)}$ are learnable parameters of the $l^{th}$ layer. 

In our case, three cascaded GCN layers ($l=3$) are employed to extract spatial-temporal feature representations of observed trajectories. Both source and target trajectories are constructed as graphs accordingly and then fed into the parameter-shared GCN layers for feature representation extraction. For simplicity, we denote the final feature representations of source trajectory domain as $F_{(s)}\in \mathbb{R}^{N^{s} \times D_f \times L_{obs}}$, and target trajectory domain as $F_{(t)}\in \mathbb{R}^{N^{t} \times D_f \times L_{obs}}$, where $N^{s}$ and $N^{t}$ are two different numbers of pedestrians from source and target domains.

% ======= Attention-Based Adaptive Learning
\subsection{Attention-Based Adaptive Learning}\label{sec:aal}
Given the misalignment of feature representations between source and target trajectory domains, we introduce an individual-wise attention-based adaptive knowledge learning module for transfer learning. Different from conventional domain adaptation situations, where each sample has determined category and fixed feature space. The feature space of trajectory sample is not fixed as the numbers of pedestrians are different in source and target trajectory domains. In order to address this misalignment problem, we propose a novel attention-based adaptive knowledge learning module to refine and effectively concentrate on the most relevant feature space for misalignment alleviation.  
 
For individual-wise attention, we first reformat the final feature representations $F_{(s)}$ and $F_{(t)}$ as,
\begin{equation}
\begin{aligned}
F_{(s)}&=\left[\textbf{f}_{(s)}^{1},\textbf{f}_{(s)}^{2},...,\textbf{f}_{(s)}^{N^{s}}\right], \quad \textbf{f}_{(s)}^{i} \in \mathbb{R}^{D_f \times L_{obs}},\\
F_{(t)}&=\left[\textbf{f}_{(t)}^{1},\textbf{f}_{(t)}^{2},...,\textbf{f}_{(t)}^{N^{t}}\right], \quad \textbf{f}_{(t)}^{i} \in \mathbb{R}^{D_f \times L_{obs}},
\end{aligned}
\end{equation}
where $\textbf{f}_{(s)}^{i}$ and $\textbf{f}_{(t)}^{i}$ correspond to the feature maps of one pedestrian from source and target trajectory domain. Then we reshape the feature maps $\textbf{f}_{(s)}^{i}$ and $\textbf{f}_{(t)}^{i}$ to the feature vector with the size of $\mathbb{R}^{D_{v}}$, where $D_{v}=D_f \times L_{obs}$.

Although the feature vector keeps the spatial-temporal information of one pedestrian, we cannot decide how representative of one pedestrian's feature vector is in one trajectory domain. Therefore, an attention module is introduced to learn the relative relevance between feature vectors and trajectory domain. The attention scores are calculated as,
\begin{equation}
\begin{aligned}
\beta_{(s)}^{i}&=\frac{\exp (\textbf{h}^{\top}\tanh(\textbf{W}_{f}\textbf{f}_{(s)}^{i}))}{\sum_{j=1}^{N^{s}}\exp(\textbf{h}^{\top}\tanh(\textbf{W}_{f}\textbf{f}_{(s)}^{j}))},\\
\beta_{(t)}^{i}&=\frac{\exp(\textbf{h}^{\top}\tanh(\textbf{W}_{f}\textbf{f}_{(t)}^{i}))}{\sum_{j=1}^{N^{t}}\exp(\textbf{h}^{\top}\tanh(\textbf{W}_{f}\textbf{f}_{(t)}^{j}))},
\end{aligned}
\label{eq5}
\end{equation}
where $\textbf{h}^{\top}$ and $\textbf{W}_{f}$ are learnable parameters. Then the final feature representations of source and target trajectory domains $\textbf{c}_{(s)}\in \mathbb{R}^{D_{v}}$ and $\textbf{c}_{(t)}\in \mathbb{R}^{D_{v}}$ are calculated as,
\begin{equation}
\begin{aligned}
\textbf{c}_{(s)}&=\sum_{i=1}^{N^s}(\beta_{(s)}^{i}\textbf{f}_{(s)}^{i}),\\
\textbf{c}_{(t)}&=\sum_{i=1}^{N^t}(\beta_{(t)}^{i}\textbf{f}_{(t)}^{i}).
\end{aligned}
\label{eq6}
\end{equation}

These two context vectors $\textbf{c}_{(s)}$ and $\textbf{c}_{(t)}$ correspond to the refined individual-level representations of source and target trajectory domains. A similarity loss $\mathcal{L}_{align}$ for distribution alignment is accordingly introduced as,
\begin{equation}
\mathcal{L}_{align}=E_{[\textbf{c}_ {(s)}\in source, \textbf{c}_{(t)}\in target ]}\left\{dist\left(\textbf{c}_{(s)},\textbf{c}_{(t)}\right)\right\}.
\label{eq7}
\end{equation}

There are multiple choices for the distance function $dist$ such as $L_2$ distance, MMD loss~\cite{Ni2013Subspace,Long2015Learning}, CORAL loss~\cite{Sun2016Deep,Zhuo2017Deep}, and adversarial loss~\cite{Ganin2014Unsupervised,2020Unsupervised}. We explore these four alignment measures in~\cref{sec:exp}, and results indicate that $L_2$ distance is more appropriate. Thus, we have,
\begin{equation}
\mathcal{L}_{align}=\frac{1}{D_f}\left|\left|\textbf{c}_{(s)}-\textbf{c}_{(t)}\right|\right|_2^2.
\label{alignloss}
\end{equation}

\subsection{Temporal Prediction Module}
Instead of making predictions frame by frame,  TCN~\cite{GCRNSM2018Shaojie} layers are employed to make future trajectory predictions based on the spatial-temporal feature representations $F_{(s)}$ from source trajectory domain. This prediction strategy is able to alleviate the error accumulating problem in sequential predictions caused by RNNs. It can also avoid gradient vanishing or reduce high computational costs~\cite{hochreiter1997long, chung2014empirical}. Recent works~\cite{mohamed2020social, shi2021sgcn} also 
utilized this strategy for prediction.

Given the feature representation $F_{(s)}\in \mathbb{R}^{N^{s} \times D_f \times L_{obs}}$, we pass  $F_{(s)}$ through TCN layers in time dimension to obtain their corresponding future trajectories. Formally, for the $l^{th}$ TCN layer, we have,
\begin{equation}
    F^{(l+1)}_{(s)}=\operatorname{TCN}(F^{(l)}_{(s)};\textbf{W}^{(l)}_{t}),
\end{equation}
where $\textbf{W}^{(l)}_{t}$ are leanable parameters of the $l^{th}$ TCN layer, $F^{(l+1)}\in \mathbb{R}^{N^{s} \times D_f \times L_{pred}}$ represents the prediction output ($L_{pred}$ represents the length to be predicted). In our case, three three cascaded TCN layers ($l=3$) are employed to obtain the final output which we refer to as $F_{(s),pred}$. 

Similar assumption is made that pedestrian coordinates $(x^{i}_{t},y^{i}_{t})$ follow a bi-variate Gaussian distribution as $(x^{i}_{t},y^{i}_{t}) \sim \mathcal{N}(\hat{\mu}^{i}_{t}, \hat{\sigma}^{i}_{t}, \hat{\rho}^{i}_{t})$, where $\hat{\mu}^{i}_{t}=(\hat{\mu}_{x},\hat{\mu}_{y})^{i}_{t}$ is the mean, $\hat{\sigma}^{i}_{t}=(\hat{\sigma}_{x},\hat{\sigma}_{y})^{i}_{t}$ is the standard deviation, and $\hat{\rho}^{i}_{t}$ is the correlation coefficient. These parameters are determined by passing $F_{(s),pred}$ through one linear layer as,
\begin{equation}
(\hat{\mu}^{i}_{t}, \hat{\sigma}^{i}_{t}, \hat{\rho}^{i}_{t}) = \operatorname{Linear}(F_{(s),pred}; \textbf{W}_{p}),
\end{equation}
where $\textbf{W}_{p}$ are learnable parameters of this linear layer.

\subsection{Objective Function}
The overall objective function consists of two terms, the prediction loss $\mathcal{L}_{pre}$ for predicting future trajectory prediction and the alignment loss $\mathcal{L}_{align}$ for aligning the distributions of source and target trajectory domains. The prediction loss $\mathcal{L}_{pre}$ is the negative log-likelihood as,
\begin{equation}
\mathcal{L}_{pre}=-\sum_{t=T_{obs}+1}^{T_{pred}} \log \left(\mathbb{P}\left((x^{i}_{t}, y^{i}_{t}) |\hat{\mu}^{i}_{t}, \hat{\sigma}^{i}_{t},  \hat{\rho}^{i}_{t}\right)\right).
\label{lpred}
\end{equation}

Note that only samples from source trajectory domain participate in the prediction phase. The whole model is trained by jointly minimizing the prediction loss $\mathcal{L}_{pre}$ and the alignment loss $\mathcal{L}_{align}$, thus we have,
\begin{equation}
\mathcal{L} = \mathcal{L}_{pre} + \lambda \mathcal{L}_{align},
\label{loss}
\end{equation}
where $\lambda$ is a hyper-parameter for balancing these two terms.

\section{Experiments}\label{sec:exp}
\begin{table*}[t]
	\centering
	\scalebox{0.63}{
		\begin{tabular}{cc|cccccccccccccccccccc|c}
			\toprule
			\multirow{2}{*}{Method} & \multirow{2}{*}{Year}&\multicolumn{20}{c|}{Performance (ADE) \quad (Source2Target)} & \multirow{2}{*}{Ave} \\
			\cmidrule(lr){3-22}
			&& A2B & A2C & A2D & A2E & B2A & B2C & B2D & B2E & C2A & C2B & C2D & C2E & D2A & D2B & D2C & D2E & E2A & E2B & E2C & E2D & \\
			\midrule

			Social-STGCNN~\cite{mohamed2020social}& 2020&
			1.83&1.58&1.30&1.31&
			3.02&1.38&2.63&1.58&
			1.16&0.70&0.82&0.54&
			1.04&1.05&0.73&0.47&
			0.98&1.09&0.74&0.50&
			1.22\\
			
			PECNet~\cite{Mangalam2020It}& 2020&
			1.97	&1.68	&1.24	&1.35	&
			3.11	&1.35	&2.69	&1.62	&
			1.39	&0.82	&0.93	&0.57	&
			1.10	&1.17	&0.92	&0.52	&
			1.01	&1.25	&0.83	&0.61   &
			1.31\\
			
			RSBG~\cite{sun2020recursive}& 2020&
			2.21	&1.59	&1.48&1.42&
			3.18	&1.49	&2.72&1.73&
			1.23	&0.87   &1.04&0.60&
			1.19    &1.21   &0.80&0.49&	
			1.09    &1.37   &1.03&0.78&
			1.38\\
			
			Tra2Tra~\cite{xu2021tra2tra}& 2021&
			1.72	&1.58	&1.27	&1.37&	
			3.32	&1.36	&2.67	&1.58&	
			1.16	&0.70	&0.85	&0.60&	
			1.09	&1.07	&0.81	&0.52&	
			1.03	&1.10	&0.75	&0.52&	
			1.25\\
			
			SGCN~\cite{shi2021sgcn}& 2021&
			1.68&	1.54&	1.26&	1.28&	
			3.22&	1.38&	2.62&	1.58&	
			1.14&	0.70&	0.82&	0.52&	
			1.05&	0.97&	0.80&	0.48&	
			0.97&	1.08&	0.75&	0.51&	
			1.22\\
			
            \midrule
			T-GNN (Ours) &  - &
			\textbf{1.13} & \textbf{1.25} & \textbf{0.94} & \textbf{1.03} & 
			\textbf{2.54} & \textbf{1.08} & \textbf{2.25} & \textbf{1.41} & 
			\textbf{0.97} & \textbf{0.54} & \textbf{0.61} & \textbf{0.23} &
			\textbf{0.88} & \textbf{0.78} & \textbf{0.59} & \textbf{0.32} & 
			\textbf{0.87 }& \textbf{0.72} & \textbf{0.65} & \textbf{0.34} & 
			\textbf{0.96}\\
			\bottomrule
	\end{tabular}}
	\vspace{-1mm}
	\caption{ADE results of our T-GNN model in comparison with existing state-of-the-art baselines on 20 tasks. ``2'' represents from source domain to target domain. A, B, C, D, and E denote ETH, HOTEL, UNIV, ZARA1, and ZARA2, respectively.}
	\vspace{-1mm}
	\label{tab:ade}
\end{table*}			

\begin{table*}[t]
	\centering
	\scalebox{0.63}{
		\begin{tabular}{cc|cccccccccccccccccccc|c}
			\toprule
			\multirow{2}{*}{Method} &\multirow{2}{*}{Year} &\multicolumn{20}{c|}{Performance (FDE) \quad (Source2Target)} & \multirow{2}{*}{Ave} \\
			\cmidrule(lr){3-22}
			&& A2B & A2C & A2D & A2E & B2A & B2C & B2D & B2E & C2A & C2B & C2D & C2E & D2A & D2B & D2C & D2E & E2A & E2B & E2C & E2D & \\
			\midrule

			Social-STGCNN~\cite{mohamed2020social}& 2020&
			3.24&2.86&2.53&2.43&
			5.16&2.51&4.86&2.88&
			2.30&1.34&1.74&1.10&
			2.21&1.99&1.41&0.88&
			2.10&2.05&1.47&1.01&
			2.30\\
			
			PECNet~\cite{Mangalam2020It}& 2020&
			3.33&	2.83&	2.53&	2.45&	
			5.23&	2.48&	4.90&	2.86&	
			2.22&	1.32&	1.68&	1.12&	
			2.20&	2.05&	1.52&	0.88&	
			2.10&	1.84&	1.45&	0.98&	
			2.29\\
			
			RSBG~\cite{sun2020recursive}& 2020&
			3.42&	2.96&	2.75&	2.50&	
			5.28&	2.59&	5.19&	3.10&	
			2.36&	1.55&	1.99&	1.37&	
			2.28&	2.22&	1.77&	0.97&	
			2.19&	2.29&	1.81&	1.34&
			2.50\\
			
			Tra2Tra~\cite{xu2021tra2tra}& 2021&
			3.29&	2.88&	2.66&	2.45&	
			5.22&	2.50&	4.89&	2.90&	
			2.29&	1.33&	1.78&	1.09&	
			2.26&	2.12&	1.63&	0.92&	
			2.18&	2.06&	1.52&	1.17&
			2.34\\

			SGCN~\cite{shi2021sgcn}& 2021&
			3.22&	2.81&	2.52&	2.40&	
			5.18&	2.47&	4.83&	2.85&	
			2.24&	1.32&	1.71&	1.03&	
			2.23&	1.90&	1.48&	0.97&	
			2.10&	1.95&	1.52&	0.99&	
			2.29\\

            \midrule
			T-GNN (Ours) & - &
            \textbf{2.18} & \textbf{2.25} & \textbf{1.78} & \textbf{1.84}&	
            \textbf{4.15} & \textbf{1.82} & \textbf{4.04} & \textbf{2.53}&	
            \textbf{1.91} & \textbf{1.12} & \textbf{1.30} & \textbf{0.87}&
            \textbf{1.92} & \textbf{1.46} & \textbf{1.25} & \textbf{0.65}&	
            \textbf{1.86} & \textbf{1.45} & \textbf{1.28} & \textbf{0.72}&
			\textbf{1.82}\\
			\bottomrule
	\end{tabular}}
		\vspace{-1mm}
	\caption{FDE results of our T-GNN model in comparison with existing state-of-the-art baselines on 20 tasks. ``2'' represents from source domain to target domain. A, B, C, D, and E denote ETH, HOTEL, UNIV, ZARA1, and ZARA2, respectively.}
	\vspace{-1mm}
	\label{tab:fde}
\end{table*}

\begin{table}[t]
	%\vspace{1mm}
	\centering
	\scalebox{0.96}{
	\begin{tabular}{c|c}
		\toprule
		\multirow{2}{*}{Method} & Average Performance \\
		\cmidrule(lr){2-2}
		& ADE/FDE \\
	    \midrule
		T-GNN+MMD~\cite{Long2015Learning} & 1.11/2.11 \\
		T-GNN+CORAL~\cite{Zhuo2017Deep} & 1.07/2.01\\
		T-GNN+GFK~\cite{2015Geodesic} & 1.15/2.08\\
		T-GNN+UDA~\cite{2020Unsupervised} & 1.07/2.09\\
		\midrule
		T-GNN (Ours) & \textbf{0.96}/\textbf{1.82}\\
		\bottomrule
    	\end{tabular}
		}
	\vspace{-1mm}
	\caption{Average performance on 20 tasks of our T-GNN model in comparison with other four commonly used DA approaches.}
	\label{tab:da}
	\vspace{-1mm}
\end{table}

\begin{table}[t]
	%\vspace{1mm}
	\centering
	\scalebox{0.96}{
	\begin{tabular}{c|ccccc}
		\toprule
		Value & $\lambda =0.01$  & $\lambda =0.1 $ & $\lambda =1 $ & $\lambda =5 $ &$\lambda =10 $\\
		\midrule
		ADE & 1.19 &1.05 &\textbf{0.96} &1.16 &1.31\\
		FDE & 2.16 &2.02 &\textbf{1.82} &2.07 &2.45\\
		\bottomrule
		\end{tabular}
		}
	\vspace{-1mm}
	\caption{Average performance on 20 tasks of our T-GNN model with 5 different values of $\lambda$.}
	\vspace{-1mm}
	\label{tab:lambda}
\end{table}

In this section, we first present the definition of our proposed new setting as well as the evaluation protocol, then we carry out extensive evaluations on our proposed T-GNN model under this new setting, in comparison with previous existing methods and different domain adaptation strategies. Additional evaluation results and feature visualizations are provided in the supplementary material.

\noindent\textbf{Datasets.} Experiments are conducted on two real-world datasets: ETH~\cite{Pellegrini2009You} and UCY~\cite{Lerner2010Crowds} as these two public datasets are widely used in this task. ETH consists of two scenes named ETH and HOTEL, and UCY consists of three scenes named UNIV, ZARA1, and ZARA2.

\noindent\textbf{Experimental Setting.}
We introduce a more general and practical setting that treats each scene as one trajectory domain. The model is trained on only one domain and tested on other four domains, respectively. Given five trajectory domains, we have total 20 trajectory prediction tasks: A $\rightarrow$ B/C/D/E, B $\rightarrow$ A/C/D/E, C $\rightarrow$ A/B/D/E, D $\rightarrow$ A/B/C/E, and E $\rightarrow$ A/B/C/D, where A, B, C, D, and E represents ETH, HOTEL, UNIV, ZARA1, and ZARA2, respectively. This setting is challenging because of the domain gap issue.

\noindent\textbf{Evaluation Protocol.}\label{sec:ep}
To ensure the fair comparison under the new setting, existing baselines are trained with one source trajectory domain as well as the validation set of the target trajectory domain. Specifically, take A $\rightarrow$ B as the example, existing baselines are trained with the training set of A and the validation set of B, then evaluated on the testing set of B. Our proposed model considers the training set of A as the source trajectory domain and the validation set of B as the target trajectory domain, then evaluated on the testing set of B. Note that the validation set and the testing set are independent of each other and there is \textbf{no overlap} sample between the validation set and the testing set. In the training phase, our proposed model only has access to the observed trajectory from the validation set.

\noindent\textbf{Baselines.} Five state-of-the-art methods are compared with our proposed method under the new setting and the evaluation protocol. \textbf{Social-STGCNN}~\cite{mohamed2020social},
\textbf{PECNet}~\cite{Mangalam2020It}, \textbf{RSBG}~\cite{sun2020recursive}, \textbf{SGCN}~\cite{shi2021sgcn}, and \textbf{Tra2Tra}~\cite{xu2021tra2tra}. We also use following four widely-used DA approaches for comparison. \textbf{T-GNN+MMD}: using the multi kernel-maximum mean discrepancies loss~\cite{Long2015Learning} as $\mathcal{L}_{align}$, \textbf{T-GNN+CORAL}: using the CORAL loss~\cite{Sun2016Deep} as $\mathcal{L}_{align}$; \textbf{T-GNN+GFK}: using the kernel-based domain adaptation strategy~\cite{2015Geodesic}, and \textbf{T-GNN+UDA}: unsupervised domain adaptive graph convolutional network using the adversarial loss~\cite{2020Unsupervised}.

\noindent\textbf{Evaluation Metrics.} Following two metrics are used to for performance evaluation. In these two metrics, $N^t$ is the total number of pedestrians in target trajectory domain, $\overline{o}^{i}_{t}$ are predictions, and $o^{i}_{t}$ are ground-truth coordinates.
\begin{itemize}
\item \textbf{Average Displacement Error (ADE):}
\begin{equation}
ADE=\frac{\sum_{i=1}^{N^t} \sum_{t=T_{obs+1}}^{T_{pred}}\|o^{i}_{t}-\overline{o}^{i}_{t}\|_{2}}{N^t\left(T_{pred}-T_{obs}\right)}.
\end{equation}
\item \textbf{Final Displacement Error (FDE):}
\begin{equation}
FDE=\frac{\sum_{i=1}^{N^t} \|o^{i}_{pred}-\overline{o}^{i}_{pred}\|_{2}}{N^t}.
\end{equation}
\end{itemize}

\noindent\textbf{Implementation Details.}
Similar with previous baselines, 8 frames are observed and the next 12 frames are predicted. The number of GCN layers is set as 3, the number of TCN layers is set as 3, and the feature dimension is set as 64. In the training phase, the batch size is set as 16 and $\lambda$ is set as 1. The whole model is trained for 200 epochs and Adam~\cite{diederik2015adam} is applied as the optimizer. We set the initial learning rate as 0.001 and change to 0.0005 after 100 epochs. In the inference phase, 20 predicted trajectories are sampled and the best amongst 20 predictions is used for evaluation.

% Quantitative Analysis
\begin{table*}[t]
	\centering
	\scalebox{0.96}{
		\begin{tabular}{cc|ccccc}
			\toprule
			\multirow{2}{*}{Variants} & \multirow{2}{*}{ID} & \multicolumn{5}{c}{Performance (ADE/FDE)} \\
			\cmidrule(lr){3-7}
			&& A2B & B2C & C2D & D2E & E2A \\
			\midrule
			T-GNN w/o GAL & 1
			& 1.51/2.34 & 1.17/1.90 & 0.69/1.42 & 0.39/0.71 & 0.90/1.98			\\  
			T-GNN w/o AAL w/ AP & 2
			& 1.78/2.85 & 1.23/2.02 & 0.77/1.53 & 0.42/0.79 & 0.96/2.03
			\\
			T-GNN w/o AAL w/ LL & 3
			& 1.81/2.91 & 1.25/2.03 & 0.76/1.48 & 0.43/0.79 & 0.94/2.01
			\\
			\midrule
			Social-STGCNN-$V_{1}$~\cite{mohamed2020social} & 4
			& 2.18/3.68 
			& 2.30/\textbf{3.21}
			&\textbf{1.59}/\textbf{2.54}
			& 1.23/1.72 
			& \textbf{1.73}/2.98\\ 
			
			SGCN-$V_{1}$~\cite{shi2021sgcn} & 5
			& \textbf{2.03}/\textbf{3.53} 
			& 2.35/3.22 
			& 1.68/2.71 
			& \textbf{1.12}/1.59 
			& 1.81/3.02\\
			
			T-GNN-$V_{1}$  & 6
			& 2.12/3.58
			& \textbf{2.28}/\textbf{3.21}
			& 1.73/2.76
			& 1.19/\textbf{1.58}
			& 1.74/\textbf{2.95}\\
			\midrule
		    Social-STGCNN~\cite{mohamed2020social} & 7
		    & 1.83/3.24
		    & 1.38/2.51
		    & 0.82/1.74
		    & 0.47/0.88
		    & 0.98/2.10
		    \\
			SGCN~\cite{shi2021sgcn} & 8
		    & 1.68/3.24
		    & 1.38/2.47
		    & 0.82/1.71
		    & 0.48/0.97
		    & 0.97/2.10
			\\
			T-GNN-$V_{2}$ & 9
			& 1.89/3.25 
			& 1.35/2.48
			& 0.88/1.93
			& 0.53/0.97
			& 0.98/2.16\\
			T-GNN (Ours) & 10
			& 1.13/2.18 & 1.08/1.82 & 0.61/1.30 & 0.32/0.65 & 0.87/1.86 \\
			\bottomrule
	\end{tabular}
	}
	\vspace{-1mm}
	\caption{Performance of different variants of T-GNN on 5 selected tasks.}
	\label{tab:abl}
	\vspace{-1mm}
\end{table*}

\subsection{Quantitative Analysis}\label{sec:qa}
\cref{tab:ade,tab:fde} show the evaluation results of 20 tasks in comparison with five existing baselines. \cref{tab:da} shows the average performance of total 20 tasks in comparison with four existing domain adaptation approaches.

\noindent\textbf{T-GNN vs Other Baselines.}
In general, our proposed T-GNN model, no matter on which task, consistently outperforms the other five baselines. Overall, our T-GNN model improves by $21.31\%$ comparing with Social-STGCNN and SGCN models on the ADE metric, and improves by $20.52\%$ comparing with PCENet and SGCN models on the FDE metric. It validates that our T-GNN model has the ability to learn transferable knowledge from source to target trajectory domain and alleviate the domain gap. As mentioned in~\cref{sec:ep}, these baselines have access to the whole validation set of the target domain while our model only has access to the observed trajectories from the validation set. Results indicate that directly training with mixed data from different trajectory domains is worse than with our domain-invariant knowledge learning approach. In addition, for tasks D2E and E2D, all the models have relatively smaller ADE and FDE. One possible reason is that domain D (ZARA1) and E (ZARA2) have similar background and surroundings, in which pedestrians may have similar moving pattern. This phenomenon further illustrates the importance of considering the domain-shift problem in trajectory prediction task.

\noindent\textbf{T-GNN vs Other DA Approaches\footnote{Performance of total 20 tasks and the implementation details of T-GNN+UDA model are provided in supplementary material since T-GNN+UDA uses an adversarial loss.}.} Generally speaking, our T-GNN model using $L_2$ distance as the alignment loss achieves the best average performance. It indicates that $L_2$ distance is more appropriate for similarity measure in trajectory prediction task. One intuitive reason is that in trajectory prediction task, high-dimensional feature representations may still reserve the spatial-level information.

\subsection{Ablation Study}\label{sec:as}
We first study the performance of different values of $\lambda$ in the objective function, and then study the contributions of each proposed component. In addition, we investigate the functionality of our proposed adaptive learning module. %We also investigate the 

\noindent\textbf{Performance Study of $\lambda$.}
The hyper-parameter $\lambda$ is used to balance the two terms in~\cref{loss}. Setting $\lambda$ too small results in the failure of alignment, on the contrary, setting $\lambda$ too large results in too heavy alignment. We set different values to find the most suitable $\lambda$. \cref{tab:lambda} shows the average performance on 20 tasks of our T-GNN model with five different values $\lambda=\left\{0.01, 0.1, 1, 5, 10\right\}$. When we set $\lambda =1$, our T-GNN model can achieve the best performance.

\noindent\textbf{Contributions of Each Component.}
We evaluate following 3 different variants of our T-GNN model on 5 selected tasks A2B, B2C, C2D, D2E, and E2A. (1) T-GNN w/o GAL denotes that the graph attention component defined in~\cref{eq1,eq2} is removed, thus $\alpha_{t;i,j}$ will not be updated during training. (2) T-GNN w/o AAL w/ AP denotes that the attention-based adaptive learning module is replaced with one average pooling layer, in which features $F_{s}$ and $F_{t}$ are reshaped and passed through one average pooling layer that operates in the ``sample'' dimension to obtain $\textbf{c}_{(s)}$ and $\textbf{c}_{(t)}$. (3) T-GNN w/o AAL w/ LL denotes that the attention-based adaptive learning module is replaced with one trainable linear layer. The results are illustrated in~\cref{tab:abl}. 

It can be observed from~\cref{tab:abl} that removing the graph attention component results in the performance reduction, which indicates the graph attention component is effective to extract relation features. Replacing our proposed attention-based adaptive learning module with either one average pooling layer or one trainable linear layer also results in the performance reduction, which indicates the effectiveness of our proposed adaptive learning module for exploring the individual-level domain-invariant knowledge.

\noindent\textbf{Effectiveness of Adaptive Learning.} Experiments are carried out to further study the effectiveness of adaptive learning module in our T-GNN model. We remove the attention-based adaptive learning module presented in~\cref{sec:aal} and disregard the alignment loss defined in~\cref{alignloss}. Thus, our model is trained only on the source trajectory domain and evaluated on one novel target trajectory domain, which we refer to as T-GNN-$V_{1}$. For further comparison, two graph-based baselines Social-STGCNN~\cite{mohamed2020social} and SGCN~\cite{shi2021sgcn} are also trained without using the validation set, which we refer to as Social-STGCNN-$V_{1}$ and SGCN-$V_{1}$. In addition, we directly train our model with mixed samples without domain-invariant adaptive learning module, which we refer to as T-GNN-$V_{2}$. The results are illustrated in~\cref{tab:abl}.

In comparison with variants 4, 5, and 6, the results indicate that the backbone of our T-GNN model is competitive with these two graph-based backbones, which validates that our T-GNN can extract effective spatial-temporal features of observed trajectories. In comparison with variants 7, 8, and 9, all three variants can achieve competitive performance since the training data is exactly the same. In addition, these three variants all outperform variants 4, 5, and 6 correspondingly, because variants 7, 8, and 9 all have access to the validation set of target trajectory domain. Results of variants 7, 8, 9 and 10 validate that our proposed domain-invariant transfer learning approach is superior to directly training with mixed data from different trajectory domains.

% ==================
% Conclusion
% ==================
\section{Conclusion}
In this paper, we delve into the domain shift challenge in the pedestrian trajectory prediction task. Specifically, a more real, practical yet challenging trajectory prediction setting is proposed. Then we propose a unified model which contains a Transferable Graph Neural Network for future trajectory prediction as well as a domain-invariant knowledge learning approach simultaneously. Extensive experiments prove the superiority of our T-GNN model in both future trajectory prediction and trajectory domain-shift alleviation. Our work is the first that studies this problem and fills the gap in benchmarks and techniques for practical pedestrian trajectory prediction across different domains.

\newpage

{\centering\large\textbf{Supplementary Material}}
\section{Overview}
In the supplementary material, we provide experimental details and more evaluation results including visualizations. We also provide our insights and discussions at the end.

\section{Experiments}
\subsection{Dataset Details}
\begin{figure*}[t]
\centering
\begin{subfigure}{0.18\textwidth}
    \centering
    \includegraphics[width=\textwidth]{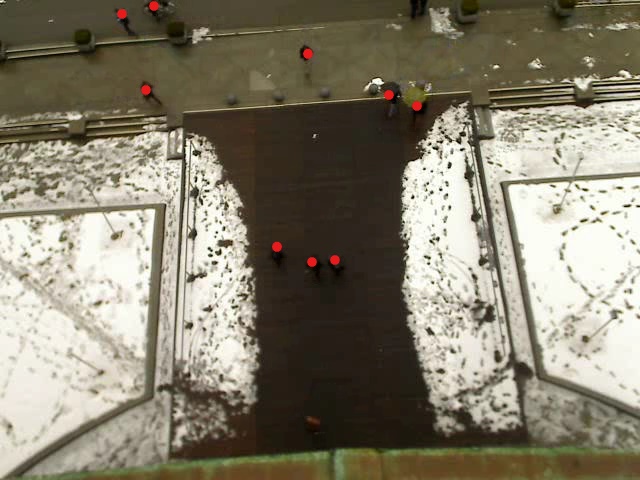}
    \caption{ETH}
\end{subfigure}
\begin{subfigure}{0.18\textwidth}
    \centering
    \includegraphics[width=\textwidth]{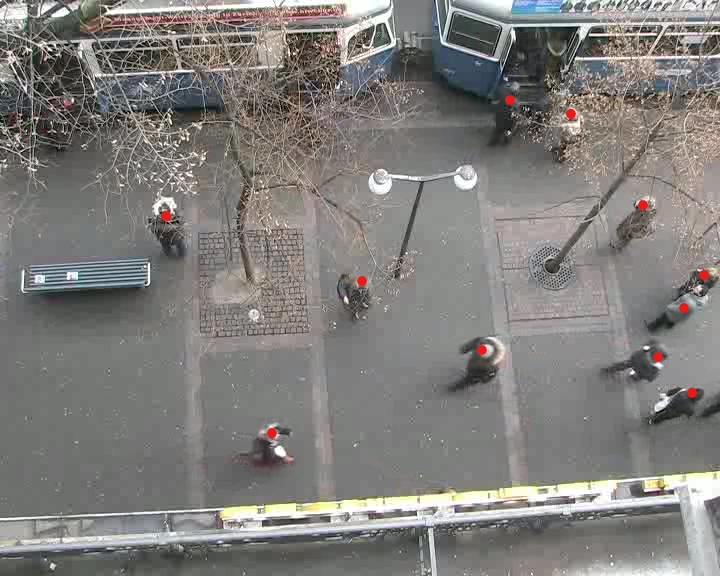}
    \caption{HOTEL}
\end{subfigure}
\begin{subfigure}{0.18\textwidth}
    \centering
    \includegraphics[width=\textwidth]{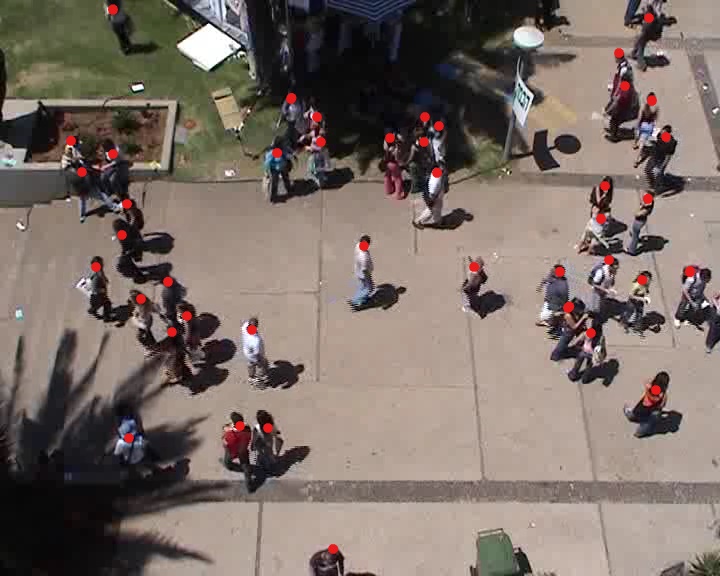}
    \caption{UNIV}
\end{subfigure}
\begin{subfigure}{0.18\textwidth}
    \centering
    \includegraphics[width=\textwidth]{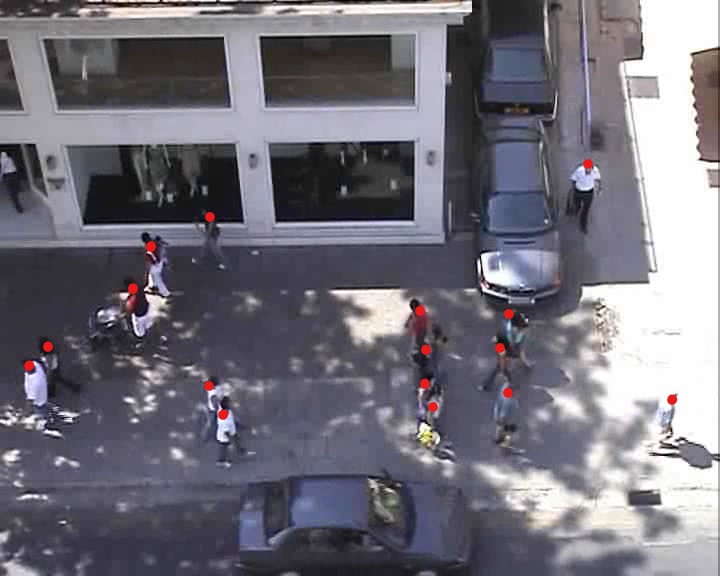}
    \caption{ZARA1}
\end{subfigure}
\begin{subfigure}{0.18\textwidth}
    \centering
    \includegraphics[width=\textwidth]{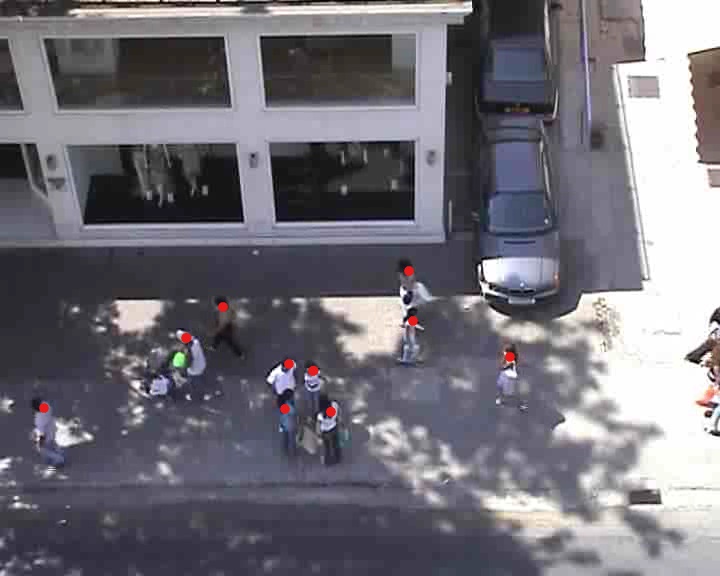}
    \caption{ZARA2}
\end{subfigure}
\caption{One frame example from five different scenes. All five scenes are from outdoor top-down view where multiple pedestrians walking in different motions (each person is denoted with one red dot). It is obvious that UNIV is much more crowded than other four scenes. In addition, ZARA1 and ZARA2 share almost the same background.}
\label{exa:fig}
\end{figure*}

There are two commonly used datasets including five scenes in our work: ETH\footnote{http://www.vision.ee.ethz.ch/en/datasets/.} ~\cite{Pellegrini2009You} and UCY\footnote{https://graphics.cs.ucy.ac.cy/research/downloads/crowd-data.}~\cite{Lerner2010Crowds}. Dataset ETH consists of two scenes: ETH and HOTEL. Dataset UCY consists of three scenes: UNIV, ZARA1, and ZARA2. Each scene contains multiple walking pedestrians with different complex walking motions. We show examples of each scene in~\cref{exa:fig} with one red dot as one person. Some basic information of two datasets is shown in~\cref{exa:table}, additional statistics of five scenes have been provided in the main body. 

\begin{table*}[t]
    \centering
    \scalebox{0.75}{
    \begin{tabular}{ccccccccc}
    \toprule
         Dataset & Year & Location & Target & Sensors & Description & Duration and tracks & Annotations & Sampling\\
         \midrule
         \textbf{ETH} & 2009 & Outdoor & People & Camera/Top-down view & Two scenes & 25 min, 650 tracks & Positions, velocities, groups, maps & @2.5Hz\\
         \textbf{UCY} & 2007 & Outdoor & People & Camera/Top-down view & Three scenes & 16.5 min, over 700 tracks & Positions, gaze directions & $-$\\
         \bottomrule
    \end{tabular}}
    \caption{Basic information ETH and UCY.}
    \label{exa:table}
\end{table*}

\subsection{Implementation Details}
We compare our proposed model with total 5 trajectory prediction baselines: Social-STGCNN~\cite{mohamed2020social},
PECNet~\cite{Mangalam2020It}, RSBG~\cite{sun2020recursive}, SGCN~\cite{shi2021sgcn}, and Tra2Tra~\cite{xu2021tra2tra}. We implemented these baselines with their provided codes: Social-STGCNN~\footnote{https://github.com/abduallahmohamed/Social-STGCNN.}, PECNet~\footnote{https://github.com/HarshayuGirase/Human-Path-Prediction.}, SGCN~\footnote{https://github.com/shuaishiliu/SGCN.}. We tried our best to reproduce the codes of RSBG and authors have shared the codes of Tra2Tra with us.We also employ 4 domain adaptation approaches: T-GNN+MMD~\cite{Long2015Learning} , T-GNN+CORAL~\cite{Sun2016Deep}, T-GNN+GFK~\cite{2015Geodesic}, and T-GNN+UDA~\cite{2020Unsupervised}. We implemented these approaches based on the codes: MMD~\footnote{https://github.com/easezyc/deep-transfer-learning.}, CORAL\footnote{https://github.com/VisionLearningGroup/CORAL.}, GFK~\footnote{https://github.com/jindongwang/transferlearning/tree/master/code/\\traditional/GFK.}, UDA~\footnote{https://github.com/GRAND-Lab/UDAGCN.}.

\subsection{Performance Study of $a_{t;i,j}$}
As mentioned in the main body, the value of $a_{t;i,j}$ in adjacency matrix $A_{t}$ is initialized as the distance between pedestrian $i$ and $j$,
\begin{equation}
    a_{t;i,j}= \|{o'}^{i}_{t} - {o'}^{j}_{t} \|_{2},
    \label{eq:a1}
\end{equation}
where $\|*\|_2$ is the $l_2$ distance, and ${o'}^{i}_{t}$ denotes the ``relative coordinates'' $({x'}^{i}_{t},{y'}^{i}_{t})$ of pedestrian $i$ at time step $t$.

As it should be other possible definitions of $a_{t;i,j}$, thus we investigate and analysis different definitions of $a_{t;i,j}$ as follows. Among these different functions, the key starting point we follow is that $a_{t;i,j}$ could be the function of the relative coordinates of pedestrians $i$ and $j$. Average ADE and FDE results are shown in~\cref{tab:suppda}.

\begin{align}
a_{t;i,j}^{reci} &=\frac{1}{\|{o'}^{i}_{t} - {o'}^{j}_{t} \|_{2}+\epsilon},\label{1}\\
a_{t;i,j}^{exp} &= \exp{\left(-\frac{\|{o'}^{i}_{t} - {o'}^{j}_{t} \|^2_2}{2\sigma^2}\right)},\\
a_{t;i,j}^{rat} &= 1- \frac{\|{o'}^{i}_{t} - {o'}^{j}_{t} \|^2_2}{\|{o'}^{i}_{t} - {o'}^{j}_{t}\|^2_2 + c}.
\label{eq:a2}
\end{align}
where $\epsilon$ and $c$ are both two small positive constants to ensure the numerical stability. In real practice, it is really difficult to have $\|{o'}^{i}_{t} - {o'}^{j}_{t}\|_2 =0$ and we set $\epsilon = c = 0.001$ though.

\begin{table}[t]
    \centering
    \begin{tabular}{c|cc}
    \toprule
         \multirow{2}{*}{Variants} & \multicolumn{2}{|c}{Performance}  \\
         \cmidrule{2-3}
         &ADE & FDE\\
          \midrule
        $a_{t;i,j}^{reci}$ & \underline{1.03} & 1.99\\
        $a_{t;i,j}^{exp}$ ($\sigma=2$) & 1.17 & 2.10\\
        $a_{t;i,j}^{exp}$ ($\sigma=4$) & 1.09 & 1.99\\
        $a_{t;i,j}^{exp}$ ($\sigma=8$) & 1.14 & 2.07\\
        $a_{t;i,j}^{rat}$ & 1.08 & \underline{1.92} \\
        \midrule
       \bm{$a_{t;i,j}$} & \textbf{0.96} & \textbf{1.82} \\
        \bottomrule
    \end{tabular}
    \caption{Average performance of total 20 tasks on ADE/FDE metric with different initializations for the adjacency matrix $A_{t}$.}
    \label{tab:suppda}
\end{table}

We can see from~\cref{tab:suppda}, the best performance comes from $a_{t;i,j}$ (\cref{eq:a1}). Function $a_{t;i,j}^{reci}$ achieves the second best performance on ADE metric and $a_{t;i,j}^{rat}$ achieves the second best performance on FDE metric, respectively.

\subsection{Results of Other DA Approaches}
\begin{table*}[ht]
	\centering
	\scalebox{0.63}{
		\begin{tabular}{cc|cccccccccccccccccccc|c}
			\toprule
			\multirow{2}{*}{Metric} & \multirow{2}{*}{Method}&\multicolumn{20}{c|}{Performance \quad (Source2Target)} & \multirow{2}{*}{Ave} \\
			\cmidrule(lr){3-22}
			&& A2B & A2C & A2D & A2E & B2A & B2C & B2D & B2E & C2A & C2B & C2D & C2E & D2A & D2B & D2C & D2E & E2A & E2B & E2C & E2D & \\
			\midrule
			\multirow{5}{*}{ADE}&
			T-GNN+MMD~\cite{Long2015Learning}
			&1.53	&1.39	&1.14	&1.19	
			&2.99	&1.18	&2.39	&1.49	
			&1.08	&0.62	&0.71	&0.42	
			&1.02	&0.89	&0.68	&0.38	
			&0.89	&0.99	&0.74	&0.41
			& 1.11 \\
			
		    &T-GNN+CORAL~\cite{Zhuo2017Deep}
		    &1.43	&1.35	&1.09	&1.12	
		    &2.87	&1.12	&2.31	&1.46	
		    &1.03	&0.58	&0.68	&0.46	
		    &0.99	&0.85	&0.66	&0.40	
		    &0.86	&0.96	&0.67	&0.41
		    & 1.07\\
		    
		    &T-GNN+GFK~\cite{2015Geodesic}
		    &1.69	&1.52	&1.20	&1.24	
		    &3.01	&1.19	&2.52	&1.55	
		    &1.11	&0.68	&0.69	&0.50	
		    &0.96	&0.89	&0.71	&0.43	
		    &0.89	&1.01	&0.75	&0.42
		    & 1.15\\
		    
		    &T-GNN+UDA~\cite{2020Unsupervised}
		    &1.41	&1.32	&0.98	&1.23	
		    &2.92	&1.20	&2.43	&1.42	
		    &1.12	&0.64	&0.62	&0.48	
		    &0.91	&0.81	&0.69	&0.35	
		    &0.91	&0.98	&0.70	&0.39
		    & 1.07\\
			
			&T-GNN (Ours) & 
			\textbf{1.13} & \textbf{1.25} & \textbf{0.94} & \textbf{1.03} & 
			\textbf{2.54} & \textbf{1.08} & \textbf{2.25} & \textbf{1.41} & 
			\textbf{0.97} & \textbf{0.54} & \textbf{0.61} & \textbf{0.23} &
			\textbf{0.88} & \textbf{0.78} & \textbf{0.59} & \textbf{0.32} & 
			\textbf{0.87 }& \textbf{0.72} & \textbf{0.65} & \textbf{0.34} & 
			\textbf{0.96}\\
			
			\midrule
			\midrule
			
			\multirow{5}{*}{FDE}
			& T-GNN+MMD~\cite{Long2015Learning}
			&2.63	&2.65	&1.98	&2.24	
			&4.86	&2.15	&4.63	&2.69	
			&2.16	&1.25	&1.52	&0.99	
			&2.20	&1.88	&1.39	&0.75	
			&2.03	&1.84	&1.46	&0.82
			& 2.11 \\
			
		    &T-GNN+CORAL~\cite{Zhuo2017Deep}
		    &2.44	&2.52	&1.82	&2.16	
		    &4.59	&1.89	&4.48	&2.68	
		    &2.09	&1.20	&1.47	&0.97	
		    &2.09	&1.83	&1.33	&0.75	
		    &2.01	&1.79	&1.38	&0.79
		    & 2.01\\
		    
		    &T-GNN+GFK~\cite{2015Geodesic}
		    &2.67	&2.66	&2.03	&2.21	
		    &4.74	&2.12	&4.88	&2.68	
		    &2.19	&1.23	&1.34	&1.01	
		    &1.96	&1.77	&1.30	&0.76	
		    &2.03	&1.83	&1.43	&0.78
		    & 2.08\\
		    
		    &T-GNN+UDA~\cite{2020Unsupervised}
		    &2.59	&2.61	&1.94	&2.25	
		    &4.81	&2.13	&4.85	&2.63	
		    &2.19	&1.29	&1.42	&1.03	
		    &2.03	&1.75	&1.37	&0.73	
		    &2.08	&1.80	&1.45	&0.76
		    & 2.09\\
			
			& T-GNN (Ours) & 
            \textbf{2.18} & \textbf{2.25} & \textbf{1.78} & \textbf{1.84}&	
            \textbf{4.15} & \textbf{1.82} & \textbf{4.04} & \textbf{2.53}&	
            \textbf{1.91} & \textbf{1.12} & \textbf{1.30} & \textbf{0.87}&
            \textbf{1.92} & \textbf{1.46} & \textbf{1.25} & \textbf{0.65}&	
            \textbf{1.86} & \textbf{1.45} & \textbf{1.28} & \textbf{0.72}&
			\textbf{1.82}\\
			\bottomrule
	\end{tabular}}
	\caption{ADE/FDE results of our T-GNN model in comparison with existing domain adaptation approaches on 20 tasks. ``2'' represents from source trajectory domain to target trajectory domain. A, B, C, D, and E denote ETH, HOTEL, UNIV, ZARA1, and ZARA2, respectively.}
	\label{tab:suppdaapp}
\end{table*}	

\cref{tab:suppdaapp} shows evaluation results of total 20 tasks when comparing with other domain adaptation approaches. For model T-GNN+UDA, in which there is an adversarial loss that measured by an extra domain classifier. One fully-connected linear layer is employed as the classifier. In specific, this kind of models needs to minimize the adversarial loss with respect to parameters of domain classifier, while maximizing it with respect to the parameters of trajectory predictor. Thus we use a a gradient reversal layer~\cite{ganin2015unsupervised} for the min-max optimization to unify the training procedure in a single step. It can be observed that our proposed adaptive learning module outperforms these domain adaptation approaches. This may show that our designed alignment loss is more appropriate for adapting fine-grained individual-level features in trajectory prediction task.

\subsection{t-SNE Visualization}
\begin{figure*}
	\centering
\begin{subfigure}{0.24\textwidth}
    \centering
    \includegraphics[width=\textwidth]{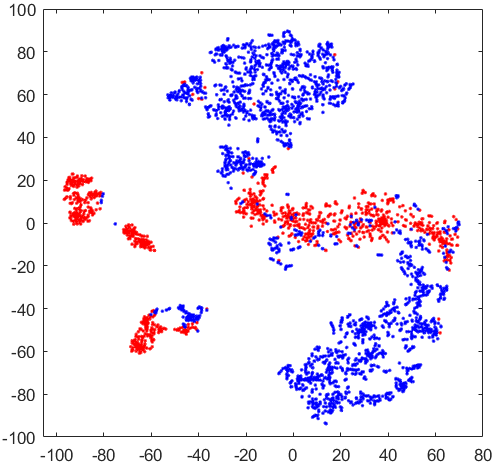}
    \caption{B$\rightarrow$D w/o AAL}
    \label{a}
\end{subfigure}
\begin{subfigure}{0.24\textwidth}
    \centering
    \includegraphics[width=\textwidth]{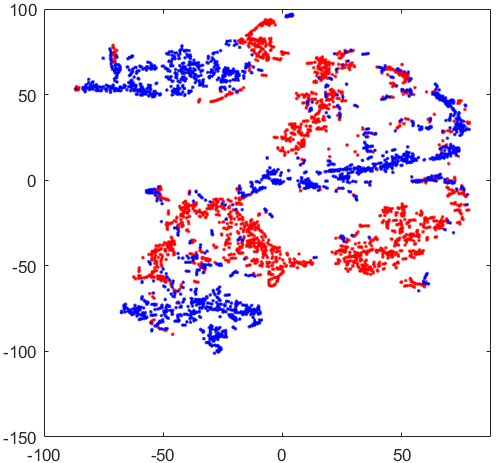}
    \caption{C$\rightarrow$E w/o AAL}
    \label{b}
\end{subfigure}
\begin{subfigure}{0.24\textwidth}
    \centering
    \includegraphics[width=\textwidth]{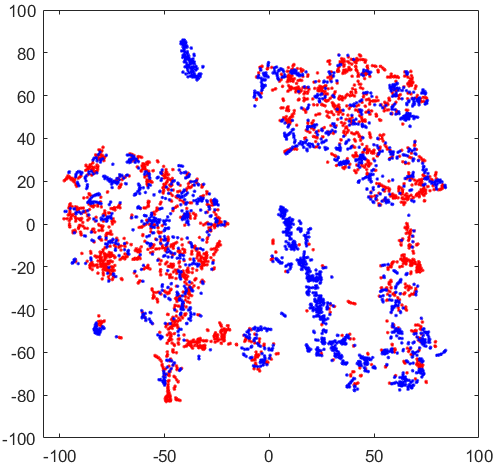}
    \caption{D$\rightarrow$E w/o AAL}
    \label{c}
\end{subfigure}	
\begin{subfigure}{0.24\textwidth}
    \centering
    \includegraphics[width=\textwidth]{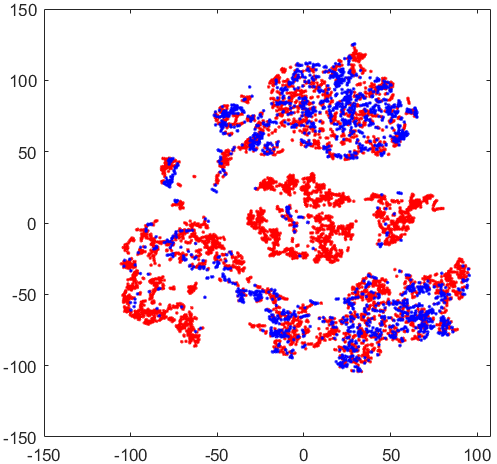}
    \caption{E$\rightarrow$D w/o AAL}
    \label{d}
\end{subfigure}		
\begin{subfigure}{0.24\textwidth}
    \centering
    \includegraphics[width=\textwidth]{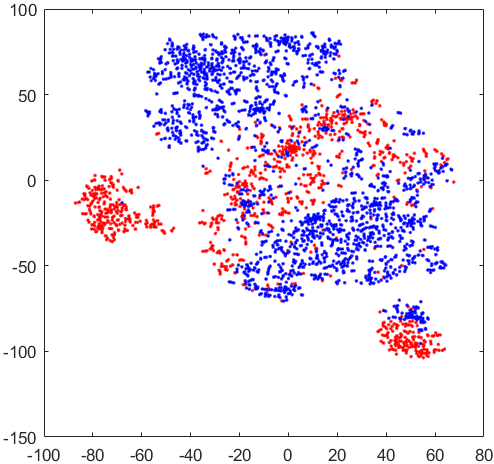}
    \caption{B$\rightarrow$D w/ AAL}
    \label{e}
\end{subfigure}		
\begin{subfigure}{0.24\textwidth}
    \centering
    \includegraphics[width=\textwidth]{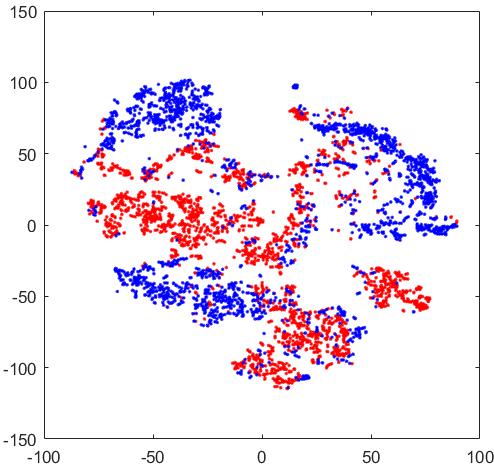}
    \caption{C$\rightarrow$E w/ AAL}
    \label{f}
\end{subfigure}		
\begin{subfigure}{0.24\textwidth}
    \centering
    \includegraphics[width=\textwidth]{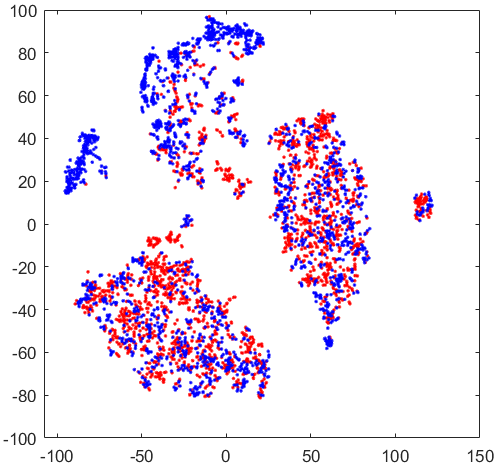}
    \caption{D$\rightarrow$E w/ AAL}
    \label{g}
\end{subfigure}	
\begin{subfigure}{0.24\textwidth}
    \centering
    \includegraphics[width=\textwidth]{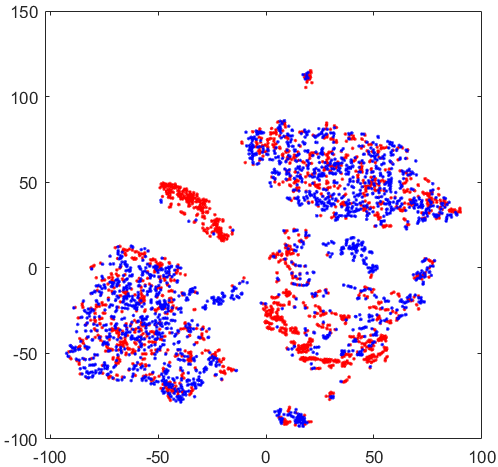}
    \caption{E$\rightarrow$D w/ AAL}
    \label{h}
\end{subfigure}	
\caption{Visualization results of the feature representations $F_{(s)}$ and $F_{(t)}$ using t-SNE. The blue and red dots denote the source and target feature representation, respectively. ``w/o AAL'' denotes that we disregard attention-based adaptive learning module(corresponding to the Variant T-GNN-$V_{2}$). ``w/ AAL'' denotes $F_{(s)}$ and $F_{(t)}$ are extracted from our proposed T-GNN model.}
\label{V}
\end{figure*}

In this section, we visualize feature representations $F_{(s)}$ and $F_{(t)}$ of the target and source trajectory domain with t-SNE~\cite{2008Visualizing} approach on of 4 tasks. \cref{V} shows the visualization examples where red and blue denote the source and target trajectory features, respectively. The first row are $F_{(s)}$ and $F_{(t)}$ without attention-based adaptive learning module , which we denote as ``w/o AAL'' (corresponding to the Variant T-GNN-$V_{2}$ in the main body). The second row are with attention-based adaptive learning module, which we denote as ``w AAL''. Each dot represents the feature of one pedestrian in the figure. Different from conventional t-SNE visualizations that applied in classification task, there is \textbf{no specific ``label''} of each dot in our task. Therefore, the cluster structure may not be clear in our task.

For task B$\rightarrow$D and C$\rightarrow$E, we can observe that features get closer with our adaptive learning module, which validates that our proposed adaptation learning module is able to alleviate the disparities across different trajectory domains. In addition, the visualization of task B$\rightarrow$D is not significant and the corresponding quantitative results of task is B$\rightarrow$D lower than others (ADE: 2.25, FDE:4.04). For task D$\rightarrow$E and E$\rightarrow$D, since D (ZARA1) and E (ZARA2) have similar scenes, we can observe from these two pairs of figures: (1) features from source and target domains have more overlaps, (2) features become more closer. It is consistent with their corresponding quantitative results (D$\rightarrow$E: 0.32/0.65, E$\rightarrow$D: 0.34/0.72). It also validates the effectiveness of our proposed adaptive learning module.

\section{Discussion}\label{sec:dis}
\noindent\textbf{Compare with general domain adaptation methods.} We delve into the domain-shift challenge in the task of pedestrian trajectory prediction in this paper. In image/video-related classification tasks, domain adaptation (DA) is a hot topic that aims to enable models to generate to novel datasets with different sample distributions. In this study, we expose the challenging domain-shift issue in future trajectory prediction. We usually consider the trajectory as two parts: observation and prediction. It is different from conventional DA tasks where data is in the form of sample-label pairs. Strictly speaking, in trajectory prediction task, the prediction part is not exactly the ``label'' of the observation part. This essential difference brings in another interesting finding that is worth exploring. In~\cref{V}, the cluster structure is not clear because there is no category of each trajectory, which means there exists distribution overlap of different trajectory domains. This kind of ``overlap'' may be reason of the variance of different tasks. If this ``overlap'' problem as well as domain-shift problem can be well-addressed simultaneously, trajectory prediction task would be more practical and promising. On the other hand, the observation and prediction parts of one trajectory are totally consistent, thus these two parts may be able to swap and supervise each other. We hope this perspective will inspire the research communities of considering the trajectory prediction problem as well as domain shift issue.

\clearpage

{\small
\balance
\bibliographystyle{ieee_fullname}
\bibliography{egbib}
}
\end{document}